%% file: iclr2026_conference.tex
\documentclass{article} 
\usepackage{iclr2026_conference,times}

\input{math_commands.tex}

\usepackage{hyperref}
\usepackage{url}
\usepackage{graphicx}
\usepackage{subfigure}
\usepackage{booktabs} 
\usepackage{bm} 
\usepackage{amsmath}
\allowdisplaybreaks 
\usepackage{float}
\usepackage{multirow}
\usepackage{makecell}

\usepackage{hyperref}
\usepackage{url}

\title{Nonparametric Variational Differential \\Privacy via Embedding Parameter Clipping}

\iclrfinalcopy

\author{%
    Dina El Zein$^{1, 2}$ \quad Shashi Kumar$^{1, 2}$ \quad James Henderson$^{1}$ \\
    $^{1}$Idiap Research Institute, Switzerland \quad
    $^{2}$EPFL, Switzerland \\
    \texttt{\{dina.el-zein, shashi.kumar, james.henderson\}@idiap.ch}
}

\newcommand{\ignore}[1]{} 
\newcommand{\bs}[1]{\boldsymbol{#1}}

\def\vx{{\bm{x}}}

\def\vz{{\bm{z}}}

\DeclareMathOperator*{\DP}{DP}

\usepackage{hyperref}
\usepackage{comment}

\usepackage{amsmath}
\usepackage{amssymb}
\usepackage{mathtools}
\usepackage{amsthm}
\usepackage[table]{xcolor}

\usepackage[capitalize,noabbrev]{cleveref}

\theoremstyle{plain}
\newtheorem{theorem}{Theorem}[section]

\theoremstyle{definition}
\newtheorem{definition}[theorem]{Definition}

\theoremstyle{remark}

\begin{document}

\maketitle

\begin{abstract}
The nonparametric variational information bottleneck (NVIB) provides the foundation for nonparametric variational differential privacy (NVDP), a framework for building privacy-preserving language models. However, the learned latent representations can drift into regions with high information content, leading to poor privacy guarantees, but also low utility due to numerical instability during training. In this work, we introduce a principled parameter clipping strategy to directly address this issue. Our method is mathematically derived from the objective of minimizing the Rényi Divergence (RD) upper bound, yielding specific, theoretically grounded constraints on the posterior mean, variance, and mixture weight parameters. We apply our technique to an NVIB based model and empirically compare it against an unconstrained baseline. Our findings demonstrate that the clipped model consistently achieves tighter RD bounds, implying stronger privacy, while simultaneously attaining higher performance on several downstream tasks. This work presents a simple yet effective method for improving the privacy-utility trade-off in variational models, making them more robust and practical.\footnote{Our code is publicly available at: \url{https://github.com/idiap/NVDP-Clipping}} 
\end{abstract}

\section{Introduction}

Large Language Models have achieved remarkable success across a wide array of complex tasks, from natural language understanding (NLU) to generation~\citep{vaswani2017attention, devlin2019bert}. This success is largely fueled by training on vast datasets, which often include sensitive user information. Consequently, these models are susceptible to privacy leakage, where they might inadvertently memorize and reveal private data~\citep{carlini2021extracting} in the embeddings shared to an adversarial third party, raising significant security and ethical concerns. Differential Privacy (DP) has emerged as the gold standard for providing formal mathematically rigorous guarantees against such leakage~\citep{dwork2006calibrating}.

While traditional DP methods often rely on direct noise injection, such as in DP-SGD~\citep{abadi2016deep}, this can severely degrade model utility. As an alternative, variational methods offer a promising path for privatizing shared text embeddings. In this framework, noise is not added directly; instead, the model learns a latent posterior distribution from which a stochastic representation is sampled to produce a sanitized embedding. Building on this idea and extending it to the attention-based representations used for text, prior work~\citep{elzein2026differential} established a framework for privacy-preserving representation learning using nonparametric variational information bottleneck (NVIB)~\citep{henderson2023vae}, called nonparametric variational differential privacy (NVDP). They demonstrated that by modeling the posterior with NVIB's flexible Dirichlet Processes, formal privacy guarantees can be derived by bounding the Rényi Divergence (RD) between this posterior and a data-independent prior, a concept inspired by the variational information bottleneck~\citep{alemi2017deep}. The resulting Bayesian Differential Privacy (BDP) guarantee~\citep{triastcyn2020bayesian} is an intrinsic part of the model's objective, resulting in a more natural integration of privacy and utility.

However, the practical application of the NVIB framework reveals a critical vulnerability: the posterior parameters are unbounded. This lack of constraint allows the parameters to drift into regions of the parameter space with high information content, resulting in a loose worst-case privacy guarantee. Furthermore, extreme parameter values can lead to numerical instability in the Rényi Divergence calculation itself. The existing NVIB framework lacks an explicit mechanism to confine the posterior to a well-behaved region that ensures both a tight privacy bound and a stable computation.

In this work, we address these limitations by introducing a principled parameter clipping strategy. Instead of relying on ad hoc heuristics, our clipping operations are mathematically derived from the objective of minimizing the RD upper bound. This analysis yields theoretically grounded constraints for the posterior mean, variance, and mixture weight parameters. We apply our method to the NVIB model and demonstrate its effectiveness on several downstream NLP tasks. Our experiments show that the clipped model achieves a superior privacy-utility trade-off, consistently yielding tighter RD bounds while improving task accuracy compared to the unconstrained NVIB baseline. Our primary contributions are:\begin{itemize}
    \item We perform a detailed analysis of the RD upper bound and derive a set of principled, theoretically grounded constraints for the posterior mean, variance, and mixture weight parameters.
    \item We implement these constraints as a novel clipping mechanism within the NVIB framework and empirically validate its effectiveness across a range of downstream NLP tasks.
    \item We demonstrate that our proposed method achieves a superior privacy-utility trade-off, yielding tighter privacy bounds while maintaining or improving task performance compared to the unconstrained baseline.
\end{itemize}

\section{Background}

This section provides the necessary background for our work. We first review the formal privacy frameworks of Rényi Differential Privacy (RDP) and BDP, which we use to quantify the privacy guarantees of our models. We then describe the architecture of the nonparametric variational differential privacy model in prior work~\citep{elzein2026differential}, which serves as the baseline for our experiments. The NVDP model is a Transformer-based architecture that incorporates a NVIB layer to learn private representations.

\subsection{Rényi Differential Privacy (RDP)}

RDP is a relaxation of standard $(\epsilon, \delta)$-DP~\citep{dwork2006calibrating} that leverages the RD~\citep{van2014renyi} to quantify the difference between probability distributions. This approach, introduced by~\citet{mironov2017renyi}, is particularly well-suited for analyzing privacy in machine learning, as it composes gracefully over multiple steps.

\begin{definition}[Rényi Divergence]\label{def:renyi}
The RD of order $\lambda > 1$ between two probability distributions $Q$ and $Q'$ over a domain $\mathcal{Z}$ is defined as:

\begin{align}
\hspace{4ex}&\hspace{-6ex}
D_\lambda(Q||Q')= \frac{1}{\lambda -1} \log\left( \int_\vz Q(\vz)\left(\frac{Q(\vz)}{Q'(\vz)}\right)^{\lambda-1}\right). 
\vspace{-1ex}
\end{align}

\end{definition}

In the context of privacy, a mechanism is said to provide $\epsilon$-RDP if for any two adjacent inputs, the divergence between their output distributions is bounded by $\epsilon$.  The order $\lambda$ determines the relative importance placed on the worst-case samples in $\mathcal{Z}$.  In our work, we adapt this concept to measure the divergence between the posterior distributions learned by our model.

\subsection{Bayesian Differential Privacy (BDP)}

While RDP provides a powerful analytical tool, BDP offers a more pragmatic interpretation of privacy, as proposed by~\citet{triastcyn2020bayesian}. BDP shifts the focus from comparing two adjacent inputs to comparing an individual's output with the expected output over the entire population. This measures how much an adversary's belief about an individual changes after observing their output, relative to a prior belief formed from the population. 

\begin{definition}[Bayesian Differential Privacy]\label{def:bdp}

A randomized mechanism $M: \mathcal{X} \rightarrow \mathcal{Z}$ satisfies $(\epsilon_\mu, \delta_\mu)$-BDP if for any individual's data point $\vx$ and a point $\vx'$ drawn from the population distribution $\mathcal{X}$, and for all measurable subsets $S \subseteq \mathcal{Z}$,:

\begin{equation}\label{eq:bdp}
    \Pr[M(\vx) \in S] \leq e^{\epsilon_\mu} \Pr[M(\vx') \in S] + \delta_\mu.
\vspace{-1ex}
\end{equation}

\end{definition}

Here, the privacy budget $\epsilon_\mu$ bounds the information gain, while $\delta_\mu$ captures the probability of failure. A key advantage of this framework is its ability to convert a bound on the RD into an interpretable $(\epsilon_\mu, \delta_\mu)$-guarantee. We detail the specific conversion mechanism used in our experiments in Section~\ref{sec:privacy_accounting}.

\subsection{Nonparametric Variational Differential Privacy (NVDP)}

To address the privacy-utility trade-off in sharing Transformer embeddings, prior work~\citep{elzein2026differential} introduced the NVDP model. The core idea of NVDP is to leverage the principles of a NVIB to learn a stochastic, information-constrained representation of an input that is sufficient for downstream tasks while being provably private. This is achieved by generating a noisy, sanitized embedding from which it is difficult to infer the original input.

\begin{figure}[ht]
\begin{center}
\centerline{\includegraphics[width=0.3\columnwidth, keepaspectratio]{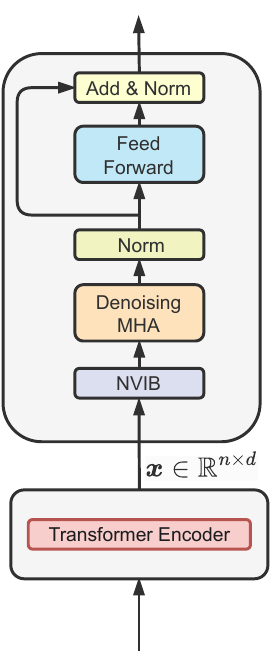}}
\vspace{-3ex}
\caption{The Nonparametric Variational Differential Privacy (NVDP) Architecture. An input sequence is first processed by a standard pretrained Transformer encoder to produce token-wise embeddings, $\boldsymbol{x} \in \mathbb{R}^{n\times d}$. These embeddings are then passed through a NVIB layer, which acts as a stochastic bottleneck by mapping them to the parameters of a posterior distribution. A sample is drawn from this distribution and processed by a Denoising Multi-Head Attention (MHA) and a Feed-Forward layer. Crucially, the standard residual skip connection around the MHA and Feed-Forward block is removed, ensuring that all information must pass through the privacy-preserving NVIB layer.}
\label{fig:NVDP}
\end{center}
\vskip -0.2in 
\end{figure}

The NVDP architecture, shown in Figure~\ref{fig:NVDP}, begins with a pretrained Transformer encoder (e.g., BERT) that produces initial embeddings. These embeddings are then passed to a single, privacy-focused Transformer block containing a NVIB layer. This NVIB layer maps the input embedding to the parameters of a posterior distribution, specifically, a set of means ($\bs{\mu}^q$), variances ($\bs{\sigma}^q$), and pseudo-counts ($\bs{\alpha}^q$) which specify a Dirichlet Process. During training, these parameters are regularized by Gaussian ($\mathcal{L}_G$) and Dirichlet ($\mathcal{L}_D$) penalty terms weighted by hyperparameters $\lambda_G$ and $\lambda_D$ respectively, which collectively enforce the information bottleneck (we set $\lambda_G = \lambda_D$). A key architectural modification for privacy is the removal of the standard residual skip connection around this block. This design choice is critical, as it forces all information to pass exclusively through the stochastic NVIB bottleneck, preventing any part of the original, unsanitized embedding from leaking to the final output.

The privacy guarantee of NVDP is realized through its sampling procedure. At both training and evaluation, we draw a noisy sample, $\bs{Z}$, from the posterior distribution defined by the learned parameters $(\bs{\mu}^q, \bs{\sigma}^q, \bs{\alpha}^q)$. This stochastic sample $\bs{Z}$, rather than the original embedding, is what is shared and used for downstream tasks. The resulting privacy leakage is quantified by adopting two complementary perspectives rooted in RDP. The first measure, RDP, ensures the indistinguishability between any two inputs by bounding the worst-case RD between their respective posterior distributions. This guarantees that an adversary cannot confidently determine which of two specific texts generated a given noisy sample. A second measure, interpreted as BDP, aggregates these pairwise divergences to quantify how much a single input stands out from the rest of the dataset. For our purposes, the worst-case RDP formulation provides the most rigorous guarantee. In the subsequent section, we will use the specific formula for this worst-case RD bound as the mathematical basis for deriving our proposed clipping strategy.

\section{Principled Clipping for Tighter Rényi Divergence Bounds}
\label{sec:clipping_method}

The core of our contribution is a principled clipping strategy designed to regularize the posterior parameters of the NVDP model. This strategy is not based on ad hoc heuristics but is derived directly from a mathematical analysis of the RD upper bound. By minimizing the terms of this bound, we can achieve tighter privacy guarantees and improve numerical stability.

\subsection{The Rényi Divergence Bound}
The RD upper bound of order $\lambda>1$ between samples from two different input embeddings ($q$ and $q'$) in the NVDP model is given by: \\ \\

\begin{align}\label{eq:rdp} 
\hspace{4ex}&\hspace{-5ex}
D_\lambda(\DP(G^q_0,\alpha^q_0)||\DP(G^{q'}_0,\alpha^{q'}_0)) 
\\ \nonumber
 \leq& -\Biggl ( \frac{1}{\lambda-1}\log\Gamma\Bigl( \lambda\alpha_{0}^q - (\lambda-1)\alpha_{0}^{q'} \Bigr)   + \log\Gamma(\alpha_{0}^{q'})-\frac{\lambda}{\lambda-1}\log\Gamma(\alpha_{0}^q) \Biggr)
 \\ \nonumber
&+~\sum_{i=1}^{n+1}\kappa_{i}\Biggl(  \frac{1}{\lambda-1}\log\Gamma(\lambda \frac{\alpha^{q}_{i}}{\kappa_{i}} - (\lambda-1)\frac{\alpha^{q'}_i}{\kappa_{i}})  + \log\Gamma(\frac{\alpha^{q'}_i}{\kappa_{i}}) -  \frac{\lambda}{\lambda-1} \log\Gamma(\frac{\alpha^{q}_{i}}{\kappa_{i}})\Biggr)
\\ \nonumber
 &+~ \sum_{i=1}^{n+1} \kappa_i \Biggl( \frac{\lambda}{2} \Biggl\lVert\frac{\bs{\mu}^q_{i} - \bs{\mu}^{q'}_i}{\bs{\sigma}'_{i}}\Biggr\rVert^2  
 + \frac{1}{1-\lambda} \bs{1} \Bigl(\log\frac{\bs{\sigma}'_{i}}{(\bs{\sigma}_{i}^{q'})^{(1-\lambda)}(\bs{\sigma}^q_{i})^\lambda}\Bigr) \Biggr), 
\vspace{-0.5ex}
\end{align}
where $ \bs{\sigma'}_{i} = \sqrt{(1{-}\lambda)(\bs{\sigma}^{q'}_i)^2 + \lambda(\bs{\sigma}^q_{i})^2}  $, $\bs{1}$ is a vector of $1$s, and $\kappa_i$ the number of vectors sampled is 1. 

\subsection{Derivation of Clipping Operations}
\label{sec:derivation}

To minimize the total divergence $D_\lambda$, we now analyze each component of the upper bound from Equation~\ref{eq:rdp} to derive a specific clipping operation for each posterior parameter. We present the high-level summary of our derivations here and provide the full step-by-step mathematical details in Appendix~\ref{app:derivations}.

\paragraph{Mean Clipping ($\bs{\mu}_i^q$).}
The first component we analyze is the mean-dependent term from Equation~\ref{eq:rdp}. For an active component $i$, this term is $D_\lambda(\bs{\mu}_i^q) = \frac{\lambda}{2} \frac{\|\bs{\mu}_i^q - \bs{\mu}_i^{q'}\|^2}{\|\bs{\sigma}'_i\|^2}$. Since the coefficient is positive, minimizing this is equivalent to minimizing the squared L2-distance $\|\bs{\mu}_i^q - \bs{\mu}_i^{q'}\|^2$. The unconstrained global minimum is achieved when $\bs{\mu}_i^q = \bs{\mu}_i^{q'}$. While forcing an exact match is too restrictive, we can introduce a budget, $2 C_\mu$, on the maximum allowable distance between two embeddings, which we enforce by constraining each individual embedding to be at most $C_\mu$ distance from the prior mean. This leads to a constrained optimization problem whose solution is to project any $\bs{\mu}_i^q$ that violates the constraint onto the L2-ball of radius $C_\mu$ centered at the prior mean. In the common case where the prior mean is zero,
this simplifies to the max-norm clipping operation:
\begin{equation}
    \hat{\bs{\mu}}_i^q = 
    \begin{cases} 
        \bs{\mu}_i^q & \text{if } \|\bs{\mu}_i^q\|_2 \le C_{\mu} \\
        C_{\mu} \frac{\bs{\mu}_i^q}{\|\bs{\mu}_i^q\|_2} & \text{if } \|\bs{\mu}_i^q\|_2 > C_{\mu}
    \end{cases}
    \label{eq:clip_mu_derived}
\end{equation}
In practice, $C_\mu$ is treated as a task-specific hyperparameter that controls the privacy-utility trade-off by limiting the information carried by the posterior mean. We select $C_\mu$ via a small grid search on the validation split and choose the smallest value that preserves near-peak validation utility while yielding the tightest worst-case RD/BDP bound. The resulting $C_\mu$ values (Appendix~\ref{app:exp_details}) vary across tasks because lower-resource tasks typically benefit from stronger regularization (smaller $C_\mu$), whereas higher-resource tasks tolerate larger budgets without degrading privacy significantly.

\paragraph{Standard Deviation Clipping ($\bs{\sigma}_i^q$).}
Next, we analyze the variance-dependent term $D_\lambda(\bs{\sigma}_i^q)= \frac{\lambda}{2} \frac{\|\bs{\mu}_i^q - \bs{\mu}_i^{q'}\|^2}{\|\bs{\sigma}'_i\|^2}
 + \frac{1}{1-\lambda} \bs{1} \left(\log\frac{\bs{\sigma'}_{i}}{(\bs{\sigma}_{i}^{q'})^{(1-\lambda)}(\bs{\sigma}^q_{i})^\lambda}\right)$. Since its leading coefficient $\frac{1}{1-\lambda}$ is negative, minimizing the term is equivalent to maximizing its inner function. By taking the derivative, we find that the theoretical optimum that minimizes the divergence is achieved when $\bs{\sigma}_i^q = \bs{\sigma}_i^{q'}$. However, a more critical requirement is ensuring the mathematical validity of the term $\bs{\sigma}'_i$, which involves a square root. For $\bs{\sigma}'_i$ to be a real number, the expression inside the square root must be non-negative, which imposes a strict lower bound on $\bs{\sigma}_i^q$:
\begin{equation}
    \sigma_{ij}^q \ge \sqrt{\frac{\lambda-1}{\lambda}}\sigma_{ij}^{q'}. \label{eq:sigma_lower_bound_main}
\end{equation}
This lower bound is the essential constraint that must be enforced, for every $\sigma_{ij}^{q'}$.  Because the NVIB regulariser pushes every $\sigma_{ij}^{q'}$ towards the prior standard deviation, and the task loss pushes it down, we know that the worst case $\sigma_{ij}^{q'}$ is bounded above by the prior standard deviation.
Therefore, our implemented strategy focuses on this aspect by clipping the posterior standard deviation from below:
\begin{equation}\label{eq:clip_sigma_derived}
    \hat{\bs{\sigma}}_i^q = \max\left(\bs{\sigma}_i^q, \sqrt{\frac{\lambda-1}{\lambda}}\bs{\sigma}_i^{q'}\right)
\end{equation}
where for $\sigma_{ij}^{q'}$ we use the prior standard deviation.
This operation guarantees the divergence is always well-defined, while not preventing the model from learning values near the theoretical optimum.

\paragraph{Alpha Clipping ($\alpha_i^q$).}
Finally, we analyze the divergence terms dependent on the pseudo-counts $\alpha_i^q$. These terms involve the log-gamma function $\log\Gamma(x)$, which is large for both large $x$ and $x$ approaching zero, and is numerically unstable for the latter. An analysis of the bound's components reveals conflicting pressures: some terms in the RD bound push for $\alpha_i^q \to \infty$ to minimize divergence, while others push for $\alpha_i^q \to 0$. This conflict makes the unconstrained minimization difficult and can lead to training failure. 

To resolve this, we constrain each $\alpha_i^q$ within a principled range $[C_{\alpha,min}, C_{\alpha,max}]$:
\begin{equation}
    \hat{\alpha}_i^q = \text{clamp}(\alpha_i^q, C_{\alpha,min}, C_{\alpha,max}).
    \label{eq:clip_alpha_derived}
\end{equation}
We set $C_{\alpha,min}$ to a small positive constant near zero to avoid the singularity of the $\log\Gamma$ function and its derivatives, which otherwise leads to unstable RD estimates and exploding gradients.
The choice of $C_{\alpha,\max}$ is motivated by both modeling and privacy considerations: $\alpha$ acts as a pseudo-count controlling the concentration of the nonparametric posterior (and thus the effective information capacity of the latent mixture), and the $\alpha$-dependent terms in Eq.~\ref{eq:rdp} grow with large pseudo-counts (via the $\log\Gamma$ combinations), which loosens the RD upper bound.
Accordingly, we choose $C_{\alpha,\max}$ to be on the order of the prior pseudo-counts (typically $\le 1$), keeping the posterior in a sparse, low-information regime consistent with the Information Bottleneck objective.
As reported in Appendix~\ref{app:exp_details}, we use smaller $C_{\alpha,\max}$ for low-resource tasks (e.g., RTE) and values closer to $1$ for higher-resource tasks (e.g., QNLI), selected from a small validation grid to balance stability, privacy, and utility. We also observe that results are not sensitive within a reasonable band around these values.

\section{Experiments}

We conduct a series of experiments to empirically evaluate our proposed clipping method. Our primary goal is to assess whether the principled clipping of posterior parameters leads to a better privacy-utility trade-off compared to an unconstrained model. We perform evaluations on both natural language understanding and speech processing tasks.

\subsection{Evaluation Metrics and Privacy Accounting}
\label{sec:privacy_accounting}

We evaluate our models on two primary axes: utility and privacy. Utility is measured using standard task-specific metrics such as Accuracy or F1-score. For privacy, we report both BDP and RD.

The RD bound is calculated for the learned distributions of our model, using the formula derived in Section~\ref{sec:clipping_method}. To offer a more concrete and interpretable privacy budget, we convert this into
a BDP guarantee. This conversion strictly follows the methodology established in prior work~\citep{elzein2026differential}, which employs the privacy accountant from~\citet{triastcyn2020bayesian}. Specifically, we leverage their Theorem 2 to derive a tight bound on the privacy loss. To quantify the privacy loss shown in Tables~\ref{tab:results_text} and ~\ref{tab:results_speech}, we fix the Rényi order to $\lambda=1.1$ (in Equation~\ref{eq:rdp}) and the BDP failure probability to $\delta_\mu=10^{-5}$ (in Equation~\ref{eq:bdp}) reporting the worst-case privacy loss across all test set pairs: specifically the maximum RD (RD max) and the corresponding BDP guarantee. 
While these tables summarize the best achieved privacy-utility trade-offs, comprehensive results for NLU tasks, across all regularization strengths, including both average and worst-case divergences, are provided in Appendix~\ref{app:text_full_results}.

\subsection{Natural Language Understanding Tasks}
\label{sec:nlp_experiments}

\paragraph{Datasets.}
We validate our approach on a diverse suite of tasks from the General Language Understanding Evaluation (GLUE) benchmark~\citep{wang2018glue}. Our evaluation includes natural language inference on RTE~\citep{dagan2005pascal} and QNLI~\citep{rajpurkar-etal-2016-squad}; sentiment analysis on SST-2~\citep{socher-etal-2013-recursive}; and paraphrase and semantic similarity detection on MRPC~\citep{dolan-brockett-2005-automatically} and STS-B~\citep{cer-etal-2017-semeval}. This selection of tasks allows us to test our model's performance on a broad range of language understanding capabilities.

\paragraph{Models and Baselines.}
To provide a comprehensive analysis, we compare three distinct models:
\begin{itemize}
    \item \textbf{Standard Classifier:} This model consists of a standard pretrained Transformer encoder with a classification head, fine-tuned directly on the downstream task. It contains no privacy-preserving mechanism and serves as an upper bound for utility.
    
    \item \textbf{NVDP (Unconstrained):} This is the model from prior work~\citep{elzein2026differential}, which applies an unconstrained NVIB layer to the Transformer embeddings. It represents the primary privacy-preserving baseline and allows us to isolate the effect of our clipping strategy.
    
    \item \textbf{NVDP-Clipped (Ours):} This is our proposed model, which incorporates the principled clipping strategy derived in Section~\ref{sec:derivation}. The parameters $(\bs{\mu}^q, \bs{\sigma}^q, \bs{\alpha}^q)$ are constrained during training to minimize the RD bound and ensure stability.
\end{itemize}

\paragraph{Implementation Details.}
For our Transformer backbones, we experiment with three widely-used architectures: BERT-Base, BERT-Large~\citep{devlin2019bert}, and RoBERTa-Base~\citep{liu2019roberta}. All models were trained using the AdamW optimizer~\citep{zhang2021revisiting, mosbach2021on}. To ensure optimal performance for each baseline, we tuned key hyperparameters, such as learning rate and batch size, for each model dataset combination. A complete breakdown of the final hyperparameters used for each experiment is provided in Appendix~\ref{app:exp_details}. 

\begin{table*}[t!]
\caption{Privacy-utility trade-off on GLUE tasks for multiple Transformer backbones. We compare our proposed \textbf{NVDP-Clipped} model against the unconstrained \textbf{NVDP} baseline for BERT-Base, BERT-Large, and RoBERTa-Base. For each model, we report its best-achieved utility score alongside privacy guarantees. Privacy is measured via BDP (BDP $\downarrow$) and the maximum (worst-case) RD (RD (max) $\downarrow$). Lower privacy values are better. The best-performing model per task is \textbf{bolded}. The non-private BERT-Base is included as a utility upper bound}
\label{tab:results_text}
\centering
\small
\begin{tabular}{@{\extracolsep{\fill}}l l r r r c r r r c r}
\toprule
& & \multicolumn{3}{c}{NVDP (Unconstrained)} & & \multicolumn{3}{c}{\textbf{NVDP-Clipped (Ours)}} & & \multicolumn{1}{c}{Non-Private} \\
\cmidrule{3-5} \cmidrule{7-9}
\textbf{Dataset} & \textbf{Metric} & \multicolumn{1}{c}{Score} & \multicolumn{1}{c}{BDP $\downarrow$} & \multicolumn{1}{c}{RD (max) $\downarrow$} & & \multicolumn{1}{c}{Score} & \multicolumn{1}{c}{BDP $\downarrow$} & \multicolumn{1}{c}{RD (max) $\downarrow$} & & \multicolumn{1}{c}{Score} \\
\midrule[1pt]
\multicolumn{11}{c}{\textbf{BERT-Base Backbone}} \\
\midrule
\multirow{2}{*}{MRPC} & Accuracy & \textbf{83.0} & 10.70 & 0.34 & & 81.6 & \textbf{10.60} & \textbf{0.013} & & 81.2 \\
 & F1 Score & \textbf{87.5} & 10.70 & 0.34 & & 87.0 & \textbf{10.60} & \textbf{0.013} & & 86.0 \\
\midrule
\multirow{2}{*}{STS-B} & Pearson & 85.2 & 20.93 & 1.41 & & \textbf{85.7} & \textbf{20.34} & \textbf{0.423} & & 86.0 \\
 & Spearman & 84.0 & 20.93 & 1.41 & & \textbf{84.7} & \textbf{20.34} & \textbf{0.423} & & 84.9 \\
\midrule
RTE & Accuracy & 64.8 & 10.90 & 1.66 & & \textbf{65.0} & \textbf{10.73} & \textbf{0.956} & & 65.9 \\
\midrule
QNLI & Accuracy & 89.5 & 12.10 & 0.75 & & \textbf{89.7} & \textbf{11.46} & \textbf{0.533} & & 89.0 \\
\midrule
SST-2 & Accuracy & \textbf{91.7} & 10.90 & 0.19 & & 91.1 & \textbf{10.80} & \textbf{0.223} & & 92.9 \\

\midrule[1pt]
\multicolumn{11}{c}{\textbf{BERT-Large Backbone}} \\
\midrule
\multirow{2}{*}{MRPC} & Accuracy & 85.0 & 10.60 & 0.611 & & \textbf{85.4} & \textbf{10.54} & \textbf{0.313} & & 84.1 \\
 & F1 Score & 89.2 & 10.60 & 0.611 & & \textbf{89.5} & \textbf{10.54} & \textbf{0.313} & & 88.5 \\
\midrule
\multirow{2}{*}{STS-B} & Pearson & 85.1 & 20.27 & 2.993 & & \textbf{86.8} & \textbf{15.93} & \textbf{0.384} & & 87.0 \\
 & Spearman & 84.9 & 20.27 & 2.993 & & \textbf{85.8} & \textbf{15.93} & \textbf{0.384} & & 86.1 \\
\midrule
RTE & Accuracy & 69.2 & 10.80 & 0.846 & & \textbf{69.7} & \textbf{10.73} & \textbf{0.109} & & 69.3 \\
\midrule
QNLI & Accuracy & \textbf{92.2} & \textbf{10.86} & 1.277 & & \textbf{92.2} & \textbf{10.86} & \textbf{1.015} & & 91.9 \\
\midrule
SST-2 & Accuracy & \textbf{94.3} & 10.48 & 0.529 & & 93.6 & \textbf{10.47} & \textbf{0.122} & & 93.3 \\

\midrule[1pt]
\multicolumn{11}{c}{\textbf{RoBERTa-Base Backbone}} \\
\midrule
\multirow{2}{*}{MRPC} & Accuracy & \textbf{87.8} & 10.55 & 0.937 & & 86.4 & \textbf{10.47} & \textbf{0.286} & & 86.9 \\
 & F1 Score & \textbf{90.8} & 10.55 & 0.937 & & 89.8 & \textbf{10.47} & \textbf{0.286} & & 90.3 \\
\midrule
\multirow{2}{*}{STS-B} & Pearson & 88.4 & 18.40 & 0.819 & & \textbf{88.6} & \textbf{15.51} & \textbf{0.245} & & 89.2 \\
 & Spearman & 87.6 & 18.40 & 0.819 & & \textbf{87.8} & \textbf{15.51} & \textbf{0.245} & & 88.6 \\
\midrule
RTE & Accuracy & 70.6 & 10.77 & 0.528 & & \textbf{71.3} & \textbf{10.55} & \textbf{0.494} & & 73.5 \\
\midrule
QNLI & Accuracy & 92.1 & 10.92 & 0.891 & & \textbf{92.2} & \textbf{10.54} & \textbf{0.118} & & 91.8 \\
\midrule
SST-2 & Accuracy & \textbf{95.1} & 10.86 & 3.007 & & 94.7 & \textbf{10.52} & \textbf{1.267} & & 94.0 \\
\bottomrule
\end{tabular}
\end{table*}
\paragraph{Results and Analysis.}
We present the primary results for the BERT-Base, BERT-Large, and RoBERTa-Base backbones in Table~\ref{tab:results_text}. This table compares the best privacy-utility trade-off achieved by the unconstrained NVDP model against our proposed NVDP-Clipped model.

The results consistently demonstrate the effectiveness of our principled clipping strategy across all model architectures. For a majority of tasks and backbones, the NVDP-Clipped model achieves a strictly superior outcome, improving both utility and privacy simultaneously. This trend is particularly pronounced on the STS-B task, where clipping consistently yields a better Pearson score while dramatically tightening the privacy budget; for BERT-Large, the BDP privacy cost improves from 20.27 to 15.93. Similarly, on tasks like RTE and QNLI, our method generally yields higher accuracy with a lower (better) privacy cost across all tested backbones. These results indicate that by constraining the posterior parameters to a stable and low-divergence region, the model is guided to learn more effective representations while incurring less privacy leakage, a benefit that scales with model size.

On other tasks, the results highlight a classic trade-off where our clipping mechanism enables a more favorable choice. For instance, across all three backbones on SST-2, the unconstrained model can sometimes achieve a fractionally higher peak accuracy, but the NVDP-Clipped model consistently provides a stronger privacy guarantee for a negligible drop in performance. This demonstrates that our clipping mechanism effectively regularizes the model, preventing it from overfitting to the last percentage point of accuracy at the cost of a looser privacy bound.

Overall, the results confirm that our formula-driven clipping successfully constrains the NVDP framework. The strategy generalizes across different model architectures and sizes, consistently leading to better or more efficient privacy-utility trade-off. The full results for all tested KL-divergence weights for each backbone are detailed in Appendix~\ref{app:text_full_results}.

\subsection{Speech Language Identification Task}
\label{sec:speech_experiments}
The goal in this task is to predict the language spoken in a given speech utterance.
\paragraph{Datasets.}
We train and evaluate on the CommonLanguage dataset~\citep{ravanelli2021speechbrain, ganesh_sinisetty_2021_5036977}, which contains approximately 45 hours of short, single-speaker read speech spanning 45 languages. All audio is resampled to a 16~kHz sampling rate. We use the dataset's default splits, comprising roughly 22k utterances for training and 6k utterances for testing.

\paragraph{Models and Baselines.}
We use a pre-trained Wav2Vec2-large model~\citep{baevski2020wav2vec} as the encoder. We attach an NVIB layer followed by a linear classification head. The Wav2Vec2 frame-level representations are passed through the NVIB layer, mean-pooled over time, and then fed to the linear head to produce a distribution over 45 language classes. We optimize a cross-entropy (CE) classification loss in addition to the losses induced by the NVIB layer. As a baseline, we consider an unconstrained NVIB variant. In our proposed NVDP-Clipped model, we constrain the variational parameters $(\bs{\mu}^q, \bs{\sigma}^q, \bs{\alpha}^q)$.

\paragraph{Implementation Details.}
We train using AdamW optimizer with a peak learning rate of $3\times 10^{-5}$. The learning rate follows a linear warmup over the first 10\% of training steps, followed by linear decay for the remaining steps. Models are trained for 10 epochs. We report the F1 score on the test set.
For NVDP (Unconstrained), we set $\lambda_G=\lambda_D=1e^{-2}$.
For NVDP-Clipped, we set $C_{\mu}=3,  C_{\alpha,max}=0.7$ and $\lambda_G=\lambda_D=1e^{-6}$. 

\paragraph{Results and Analysis.}
The results in Table~\ref{tab:results_speech} demonstrate that the benefits of our principled clipping strategy extend beyond text to speech embeddings. While the unconstrained NVDP baseline achieves better F1 score, it does so by allowing the posterior to drift into regions that incur a significantly higher privacy cost and a looser RD bound.
In contrast, the NVDP-Clipped model achieves a more favorable privacy-utility trade-off. It provides a substantially tighter privacy guarantee and a lower worst-case divergence while maintaining competitive task performance. This confirms that the constraints derived in Section~\ref{sec:derivation} are robust across different modalities and architectures (Wav2Vec2 vs. BERT), successfully preventing the model from sacrificing formal privacy for marginal gains in utility.
\begin{table}[t]
\caption{Privacy-utility trade-off for Speech Language Identification using a Wav2Vec2 backbone. Consistent with the NLP results, the clipped model achieves a tighter privacy profile (lower BDP and RD) with minimal impact on utility measure by F1 score.}
    \label{tab:results_speech}
    \centering
    \small
    \begin{tabular}{l c c c}
    \toprule
    \textbf{Model} & \textbf{F1}$\uparrow$ & \textbf{BDP}$\downarrow$ & \textbf{RD} (max) $\downarrow$ \\
    \midrule
       NVDP (Unconstrained) & \textbf{83.7} & 10.46 & 4.044 \\
       NVDP-Clipped & 82.6 & \textbf{9.90} & \textbf{3.567} \\
    \bottomrule
    \end{tabular}
\end{table}

\section{Conclusion}
In this work, we introduce a critical enhancement to the NVDP framework to provide stronger and more reliable privacy guarantees. We addresse the limitation of unbounded worst-case privacy loss in the original method by proposing a principled clipping strategy. Our approach enforces explicit bounds on all three key latent parameters that define the posterior distribution: the L2 norm of the latent mean $ \hat{\bs{\mu}}_i^q $, the magnitude of the latent variance $ \hat{\bs{\sigma}}_i^q $, and the value of the pseudo count $\hat{\alpha}_i^q$. By constraining these parameters, our NVDP-Clipped model ensures that the RD is provably bounded. Our experiments confirm that this method significantly reduces the worst-case privacy loss, 
while preserving high utility on downstream NLP tasks, thereby transforming NVDP into a more practical and trustworthy tool for real-world applications.

\subsection*{Acknowledgments}
We would like to thank other members of Idiap and its NLU group for helpful comments and suggestions.  
This research was funded in part by the Swiss National Science Foundation (SNSF) under grant number 10003729.

\bibliography{iclr2026_conference}
\bibliographystyle{iclr2026_conference}
\newpage

\appendix

\section{Detailed Derivations of Clipping Operations}
\label{app:derivations}

In this section, we provide the full step-by-step derivation for each of the clipping operations summarized in Section~\ref{sec:derivation}.

\subsection{Mean Clipping \texorpdfstring{($\bs{\mu}_i^q$)}{(mu\_i\^q)}}
\paragraph{Derivation.}
We seek to minimize the part of the formula that depends on $\bs{\mu}_i^q$. This term appears only in the final summation of Equation \ref{eq:rdp}. For a component $i$, this term is:
\begin{equation}
    D_\lambda(\bs{\mu}_i^q) = \frac{\lambda}{2} \frac{\|\bs{\mu}_i^q - \bs{\mu}_i^{q'}\|^2}{\|\bs{\sigma}'_i\|^2}.
\end{equation}
The minimization problem with respect to $\bs{\mu}_i^q$ is:
\begin{equation}
    \min_{\bs{\mu}_i^q} D_\lambda(\bs{\mu}_i^q) = \min_{\bs{\mu}_i^q} \left( \frac{\lambda}{2\|\bs{\sigma}'_i\|^2} \cdot \|\bs{\mu}_i^q - \bs{\mu}_i^{q'}\|^2 \right).
\end{equation}
Since the coefficient $\frac{\lambda}{2\|\bs{\sigma}'_i\|^2}$ is positive, minimizing the expression is equivalent to minimizing the squared L2-distance between $\bs{\mu}_i^q$ and $\bs{\mu}_i^{q'}$. The unconstrained global minimum is achieved when $\bs{\mu}_i^q = \bs{\mu}_i^{q'}$.


We therefore introduce a budget, $2 C_\mu$, on the maximum allowable distance from $\bs{\mu}_i^{q'}$, which we enforce by constraining each individual embedding mean (including $\bs{\mu}_i^{q}$ and $\bs{\mu}_i^{q'}$) to be at most $C_\mu$ distance from the prior mean.
The solution to this constrained optimization problem is to project any $\bs{\mu}_i^q$ whose distance from the prior mean violates the constraint onto the boundary of the feasible region (the L2-ball of radius $C_\mu$ centered at the prior mean). This leads to the following operation:
\begin{align}
    \text{If } \|\bs{\mu}_i^q - \bs{\mu}_i^{q'}\|_2 > C_\mu, \quad \text{then set } 
    \quad \hat{\bs{\mu}}_i^q = \bs{\mu}_i^{q'} + C_\mu \frac{\bs{\mu}_i^q - \bs{\mu}_i^{q'}}{\|\bs{\mu}_i^q - \bs{\mu}_i^{q'}\|_2}.
\end{align}
where for $\bs{\mu}_i^{q'}$ we use the prior mean.
In the common case where the prior mean is zero, this simplifies to the max-norm clipping operation, as presented in Equation~\ref{eq:clip_mu_derived}.

\subsection{Standard Deviation Clipping (\texorpdfstring{$\bs{\sigma}_i^q$}{sigma\_i\^q})}

\paragraph{Derivation.}
We seek to minimize the component of the RD that depends on the posterior standard deviation, $\bs{\sigma}_i^q$. For a single component $i$, this full variance-dependent term is:
\begin{align}
    D_\lambda(\bs{\sigma}_i^q) &= \frac{\lambda}{2} \frac{\|\bs{\mu}_i^q - \bs{\mu}_i^{q'}\|^2}{\|\bs{\sigma}'_i\|^2} 
    + \frac{1}{1-\lambda} \bs{1} \left(\log\frac{\bs{\sigma'}_{i}}{(\bs{\sigma}_{i}^{q'})^{(1-\lambda)}(\bs{\sigma}^q_{i})^\lambda}\right),
\end{align}
where $\bs{\sigma}'_i = \sqrt{(1-\lambda)(\bs{\sigma}_i^{q'})^2 + \lambda(\bs{\sigma}_i^q)^2}$. Our goal is to find the value of $\bs{\sigma}_i^q$ that minimizes this expression.

First, for the RD to be well-defined, the term $\bs{\sigma}'_i$ must be a real number. This requires the expression inside its defining square root to be non-negative. For each dimension $j$, this imposes a hard constraint on the value of $\sigma_{ij}^q$:
\begin{align}
    (1-\lambda)(\sigma_{ij}^{q'})^2 + \lambda(\sigma_{ij}^q)^2 &\ge 0 \\
    \lambda(\sigma_{ij}^q)^2 &\ge (\lambda-1)(\sigma_{ij}^{q'})^2 \quad (\text{since } \lambda > 1) \\
    \sigma_{ij}^q &\ge \sqrt{\frac{\lambda-1}{\lambda}}\sigma_{ij}^{q'} \label{eq:sigma_lower_bound}
\end{align}
This inequality defines a strict lower bound on the posterior standard deviation, establishing the feasible region for our optimization. If any dimension of $\bs{\sigma}_i^q$ falls below this value, the divergence becomes mathematically undefined.

To express this constraint on any pair of embeddings as a constraint on any individual embedding, we take advantage of the fact that every $\sigma_{ij}^{q'}$ is bounded above by the prior standard deviation.  This is because the task loss always pushes $\sigma_{ij}^{q'}$ down towards zero so as to reduce the noise and make the embedding more informative for performing the task.  The only loss pushing $\sigma_{ij}^{q'}$ up is the NVIB regulariser, which pushes it towards the prior standard deviation.  So any trained $\sigma_{ij}^{q'}$ will be between zero and the prior standard deviation.  This allows us to use the prior standard deviation as a bound on the worst case $\sigma_{ij}^{q'}$ in the above constraint on $\sigma_{ij}^{q}$.

We therefore propose a clipping strategy that directly enforces this validity constraint. The operation, applied element-wise to the model's output $\bs{\sigma}_i^q$, is:
\begin{equation} \label{eq:sigma_clip_operation}
    \hat{\bs{\sigma}}_i^q = \max\left(\bs{\sigma}_i^q, \sqrt{\frac{\lambda-1}{\lambda}}\bs{\sigma}_i^{q'}\right).
\end{equation}
where for $\bs{\sigma}_i^{q'}$ we use the prior standard deviation.

We can now justify this operation as the correct solution to the constrained minimization problem. By inspection, the divergence term $D_\lambda(\bs{\sigma}_i^q)$ is minimized when the posterior distribution most closely matches the alternative posterior distribution, i.e., when $\bs{\sigma}_i^q = \bs{\sigma}_i^{q'}$. At this point, the log-based contribution to the divergence becomes zero. Since $\sqrt{(\lambda-1)/\lambda} < 1$, this unconstrained optimum $\bs{\sigma}_i^{q'}$ always lies within the feasible region defined by the lower bound in Equation~\ref{eq:sigma_lower_bound}.

The problem is thus to find the point in the feasible region that is closest to the true optimum $\bs{\sigma}_i^{q'}$. If the model's output $\bs{\sigma}_i^q$ is already in this region, no change is needed. If it violates the constraint (i.e., it is smaller than the lower bound), the optimal solution is to project it to the closest point in the feasible region, which is the boundary itself. This is precisely what the `max` operation in Equation~\ref{eq:sigma_clip_operation} achieves.

\subsection{Alpha Clipping \texorpdfstring{($\alpha_i^q$)}{(alpha\_i\^q)}}

\paragraph{Derivation.}
The divergence terms dependent on the pseudo-counts $\alpha_i^q$ and their sum $\alpha_0^q = \sum_i \alpha_i^q$, which we denote as $D_\lambda(\bs{\alpha}^q)$, are composed of both global and local parts:
\begin{align}
    D_\lambda(\bs{\alpha}^q) = 
    &-\Biggl ( \frac{1}{\lambda-1}\log\Gamma\Bigl( \lambda\alpha_{0}^q - (\lambda-1)\alpha_{0}^{q'} \Bigr) \label{eq:alpha_global_appendix}
    + \log\Gamma(\alpha_{0}^{q'}) -\frac{\lambda}{\lambda-1}\log\Gamma(\alpha_{0}^q) \Biggr)  \\
    &+~\sum_{i=1}^{n+1} \Biggl(  \frac{1}{\lambda-1}\log\Gamma(\lambda \alpha^{q}_{i} - (\lambda{-}1)\alpha^{q'}_i) \label{eq:alpha_local_appendix}
    + \log\Gamma(\alpha^{q'}_i) -  \frac{\lambda}{\lambda{-}1} \log\Gamma(\alpha^{q}_{i})\Biggr).
\end{align}

We know of no closed-form formula for minimization of this expression with respect to $\bs{\alpha}^q$. This is due to the complex nature of the log-gamma function, $\log\Gamma(x)$, and, more importantly, because different components of the formula create conflicting pressures on the values of $\alpha_i^q$.  The log-gamma function $\log\Gamma(x)$ is large both for large values of $x$ and as $x$ approaches zero, and has a singularity at zero.  We can reorganize this formula into three pairs of terms (aligned vertically in \eqref{eq:alpha_global_appendix} and \eqref{eq:alpha_local_appendix}) consisting of one global term and one local term, where each pair has the form $\sum_i \log\Gamma(x_i) - \log\Gamma(\sum_i x_i)$ (or its negative).  This function is large when any $x_i$ approaches zero and otherwise decreases as $\sum_i x_i$ grows.  

First we consider the positive instance of this function in the terms 
$\sum_{i=1}^{n+1} \log\Gamma(\alpha^{q'}_i) - \log\Gamma(\alpha_{0}^{q'})$.
To keep this instance from exploding, we need to ensure that each individual $\alpha^{q'}_i$ does not get too close to zero.  We therefore constrain all $\alpha^{q}_i$ to be above a small value 
$C_{\alpha,min}$.  
This constraint also helps prevent the other positive instance of this function,
$\frac{1}{\lambda-1}\left( \sum_{i=1}^{n+1} \log\Gamma\left(\lambda \alpha^{q}_{i} - (\lambda{-}1)\alpha^{q'}_i\right)
- \log\Gamma\left( \lambda\alpha_{0}^q - (\lambda-1)\alpha_{0}^{q'} \right) \right)$,
from exploding.

Next we consider the negative instance of this function in the terms 
$-\frac{\lambda}{\lambda{-}1} \left( \sum_{i=1}^{n+1}  \log\Gamma(\alpha^{q}_{i}) - \log\Gamma(\alpha_{0}^q) \right)$.  
Since the weight $-\frac{\lambda}{\lambda{-}1}$ is always negative, we want to keep $\sum_{i=1}^{n+1}  \log\Gamma(\alpha^{q}_{i}) - \log\Gamma(\alpha_{0}^q)$ from getting too small.  This implies preventing $\sum_i \alpha_{i}^{q} = \alpha_{0}^q$ from getting too big, which we do by constraining the individual $\alpha_{i}^{q}$ to be below a value 
$C_{\alpha,max}$.

\ignore{***
To understand this instability, we first analyze the local term, Equation~\ref{eq:alpha_local_appendix}, for a single active component $i$:
\begin{itemize}
    \item \textbf{Term 1: $-\frac{\lambda}{\lambda-1}\log\Gamma(\alpha_i^q)$.} For $\lambda > 1$, the coefficient $-\frac{\lambda}{\lambda-1}$ is negative. Since $\log\Gamma(x)$ is a monotonically increasing function for $x > 1.5$, this term moves towards $-\infty$ as $\alpha_i^q$ grows. To minimize the divergence, this term pushes $\alpha_i^q$ towards infinity.

    \item \textbf{Term 2: $+\frac{1}{\lambda-1}\log\Gamma(\lambda \alpha^{q}_{i} - (\lambda{-}1)\alpha^{q'}_i)$.} The coefficient $\frac{1}{\lambda-1}$ is positive. The argument of the function, $\lambda \alpha^{q}_{i} - (\lambda{-}1)\alpha^{q'}_i$, also grows with $\alpha_i^q$. This term therefore moves towards $+\infty$ as $\alpha_i^q$ grows. To minimize the divergence, this term pushes $\alpha_i^q$ to be as small as possible.
\end{itemize}
This direct conflict—where different components push $\alpha_i^q$ towards opposite extremes—makes the unconstrained optimization problem difficult. An unbounded $\alpha_i^q$ would cause the divergence to explode due to the second term. Furthermore, the log-gamma function is undefined for non-positive values and singular as its argument approaches zero, creating numerical instability for small $\alpha_i^q$.

This instability at the local level for each $\alpha_i^q$ naturally extends to the global term in Equation \ref{eq:alpha_global_appendix}, which depends on their sum $\alpha_0^q$. 
***}

Thus, to enforce these constraints on $D_\lambda(\bs{\alpha}^q)$ and to ensure a stable training process, we introduce a clipping operation at the local level. We constrain each individual $\alpha_i^q$ to lie within a "safe" range $[C_{\alpha,min}, C_{\alpha,max}]$, where $C_{\alpha,min} > 0$ to avoid the singularity at zero. This operation not only helps to constrain $D_\lambda(\bs{\alpha}^q)$, as argued above, but is also a practical necessity for numerical stability. By constraining each $\alpha_i^q$, we inherently also constrain their sum, ${\alpha}_0^q = \sum_i {\alpha}_i^q$, thereby stabilizing both the positive and the negative instances of the function $\sum_i \log\Gamma(x_i) - \log\Gamma(\sum_i x_i)$ in the divergence.

\section{Experimental Details and Full Results for NLU Tasks}
\subsection{Hyperparameter Details}\label{app:exp_details}

All models for the NLU tasks were fine-tuned on their respective GLUE tasks. The general training hyperparameters for each backbone are listed in Table~\ref{tab:hyperparams_general}. Our clipping strategy introduces two task-specific hyperparameters: $\hat{\bs{\mu}}_i^q$, which defines the maximum L2 norm for the latent means, and $\hat{\alpha}_i^q$, which sets the maximum value for the pseudo-counts. The values used for these clipping hyperparameters are detailed in Table~\ref{tab:hyperparams_clipping}.

\begin{table}[h!]
\caption{General training hyperparameters for each model backbone.}
\label{tab:hyperparams_general}
\centering
\small
\begin{tabular}{@{}lccc@{}}
\toprule
\textbf{Hyperparameter} & \textbf{BERT-Base} & \textbf{BERT-Large} & \textbf{RoBERTa-Base} \\
\midrule
Learning Rate & \texttt{7e-5} & \texttt{2e-5} & \texttt{2e-5} \\
Warmup Ratio & \texttt{0.06} & \texttt{0.1} & \texttt{0.06} \\
Adam Epsilon & \texttt{1e-8} & \texttt{1e-8} & \texttt{1e-8} \\
Train Batch Size & \texttt{32} & \texttt{32} & \texttt{32} \\
Eval Batch Size & \texttt{16} & \texttt{16} & \texttt{16} \\
Max Sequence Length & \texttt{512}/ \texttt{128} & \texttt{512} / \texttt{128} & \texttt{512} / \texttt{128} \\
\bottomrule
\end{tabular}
\end{table}
\begin{table}[h!]
\caption{Task-specific clipping hyperparameters reported as the triple ($C_{\mu}, C_{\alpha,min}, C_{\alpha,max}$). These constraints ensure numerical stability and minimize the RD upper bound.}
\label{tab:hyperparams_clipping}
\centering
\small
\setlength{\tabcolsep}{8pt} 
\begin{tabular}{@{}l ccc@{}}
\toprule
\textbf{Task} & \textbf{BERT-Base} & \textbf{BERT-Large} & \textbf{RoBERTa-Base} \\
\midrule
MRPC  & (2, 0, 0.5)  & (3, 0, 0.7)  & (3, 0, 0.7)  \\
QNLI  & (7, 0, 1)  & (7, 0, 1)  & (7, 0, 1)  \\
SST-2 & (2, 0, 1)  & (3, 0, 1)  & (3, 0, 1)  \\
RTE   & (6, 0, 0.5)  & (6, 0, 0.6)  & (6, 0, 0.6)  \\
STS-B & (10, 0, 1) & (10, 0, 1) & (10, 0, 1) \\
\bottomrule
\end{tabular}
\end{table}

\subsection{Full Results Tables}\label{app:text_full_results}

This section provides the comprehensive experimental results for all model backbones for NLU tasks presented in Section~\ref{sec:nlp_experiments}. The results for BERT-Base, BERT-Large, and RoBERTa-Base are detailed in Table~\ref{tab:appendix_text_full_results_bert_base}, Table~\ref{tab:appendix_text_full_results_bert_large}, and Table~\ref{tab:appendix_text_full_results_roberta_base}, respectively. Each table presents the full privacy-utility trade-off across varying strengths of the training regularization weights ($\lambda_{G}=\lambda_{D}$). These hyperparameters control the the Gaussian and Dirichlet information bottlenecks of the information bottleneck, respectively, allowing us to explore the Pareto frontier between task performance and privacy-preserving noise levels.

\begin{table*}[t!]
\caption{Full experimental results for \textbf{BERT-Base} on GLUE tasks. We compare the unconstrained and clipped NVDP models across different strengths of the KL-divergence regularization weight ($\lambda_{G}=\lambda_{D}$). For each model, we report Utility (task-specific scores) and privacy metrics BDP ($\epsilon_\mu$) and RD. The RD metric is reported as max/avg, representing the worst-case and average-case divergence across all example pairs, respectively. Lower values are better for all privacy metrics ($\downarrow$).}
\label{tab:appendix_text_full_results_bert_base}
\centering
\small
\resizebox{\textwidth}{!}{%
\begin{tabular}{@{} l l c c c c c @{}}
\toprule
\textbf{Model} & \textbf{Metric} & \multicolumn{1}{c}{MRPC} & \multicolumn{1}{c}{STS-B} & \multicolumn{1}{c}{RTE} & \multicolumn{1}{c}{QNLI} & \multicolumn{1}{c}{SST-2} \\
\cmidrule(l){3-7}
& & Acc / F1 & Pearson / Spearman & Accuracy & Accuracy & Accuracy \\
\midrule
\multirow{2}{*}{BERT$_\text{Base}$ (Non-Private)} & Utility & 81.2 / 86.0 & 86.0 / 84.9 & 65.9 & 89.0 & 92.9 \\
& Privacy & - & - & - & - & - \\
\midrule
\multicolumn{7}{l}{\textbf{NVDP (Unconstrained)}} \\
\midrule
\multirow{3}{*}{$\lambda=1e-3$} & Utility & 82.5 / 87.1 & 85.2 / 84.0 & 64.8 & 89.5 & 91.7 \\
& BDP ($\epsilon_\mu$) $\downarrow$ & 10.95 & 20.93 & 10.90 & 12.10 & 11.19 \\
& RD (max/avg) $\downarrow$ & 0.89 / 0.06 & 1.41 / 0.10 & 1.66 / 0.12 & 0.75 / 0.06 & 1.00 / 0.06 \\
\cmidrule(l){2-7}
\multirow{3}{*}{$\lambda=1e-2$} & Utility & 83.0 / 87.5 & 84.6 / 83.6 & 61.3 & 89.2 & 91.7 \\
& BDP ($\epsilon_\mu$) $\downarrow$ & 10.70 & 20.77 & 10.70 & 11.30 & 10.90 \\
& RD (max/avg) $\downarrow$ & 0.34 / 0.02 & 1.22 / 0.05 & 0.87 / 0.04 & 0.71 / 0.09 & 0.19 / 0.01 \\
\cmidrule(l){2-7}
\multirow{3}{*}{$\lambda=1e-1$} & Utility & 68.3 / 80.2 & 78.4 / 77.7 & 50.3 & 49.5 & 90.7 \\
& BDP ($\epsilon_\mu$) $\downarrow$ & 10.50 & 20.38 & 10.60 & 10.48 & 10.70 \\
& RD (max/avg) $\downarrow$ & 0.04 / 0.01 & 0.22 / 0.01 & 0.10 / 0.01 & 0.016 / 0.003 & 0.016 / 0.004 \\
\cmidrule(l){2-7}
\multirow{3}{*}{$\lambda=1$} & Utility & 66.5 / 79.9 & 82.7 / 82.9 & 50.3 & 49.5 & 49.9 \\
& BDP ($\epsilon_\mu$) $\downarrow$ & 10.40 & 18.65 & 10.40 & 10.46 & 10.40 \\
& RD (max/avg) $\downarrow$ & 0.008 / 0.002 & 0.03 / 0.004 & 0.005 / 0 & 0.007 / 0.001 & 0.01 / 0.002 \\
\midrule
\multicolumn{7}{l}{\textbf{NVDP-Clipped (Ours)}} \\
\midrule
\multirow{3}{*}{$\lambda=1e-3$} & Utility & 80.6 / 85.8 & 85.7 / 84.7 & 65.0 & 89.7 & 91.1 \\
& BDP ($\epsilon_\mu$) $\downarrow$ & 10.80 & 20.34 & 10.73 & 11.46 & 10.80 \\
& RD (max/avg) $\downarrow$ & 0.112/ 0.007 & 0.423 / 0.036 & 0.956 / 0.075 & 0.533 / 0.037 & 0.223 / 0.035 \\
\cmidrule(l){2-7}
\multirow{3}{*}{$\lambda=1e-2$} & Utility & 80.7 / 85.3 & 85.5 / 84.5 & 64.8 & 88.8 & 89.9 \\
& BDP ($\epsilon_\mu$) $\downarrow$ & 10.60 & 19.85 & 10.70 & 11.18 & 10.60 \\
& RD (max/avg) $\downarrow$ & 0.130 / 0.011 & 0.410 / 0.036 & 0.100 / 0.017 & 0.310 / 0.017 & 0.090 / 0.010 \\
\cmidrule(l){2-7}
\multirow{3}{*}{$\lambda=1e-1$} & Utility & {81.6} / {87.0} & {85.6} / {84.5} & {64.9} & {88.2} & {89.7} \\

& BDP ($\epsilon_\mu$) $\downarrow$ & {10.60} & {19.70} & {10.70} & {10.47} & {10.60} \\
& RD (max/avg) $\downarrow$ & {0.013} / {0.002} & {0.114} / {0.011} & {0.080} / {0.006} & {0.004} / {0.000} & {0.020} / {0.003} \\
\cmidrule(l){2-7}
\multirow{3}{*}{$\lambda=1$} & Utility & {66.4} / {79.8} & {85.1} / {83.8} & {49.3} & {49.5} & {49.8} \\
& BDP ($\epsilon_\mu$) $\downarrow$ & {10.40} & {19.01} & {10.40} & {10.46} & {10.40} \\
& RD (max/avg) $\downarrow$ & {0} / {0} & {0.051} / {0.007} & {0.001} / {0} & {0.004} / {0} & {0.003} / {0.001} \\
\bottomrule
\end{tabular}
}
\end{table*}

\begin{table*}[t!]
\caption{Full experimental results for \textbf{BERT-Large} on GLUE tasks, showing the complete privacy-utility trade-off across all tested KL-divergence regularization weights ($\lambda_{G}, \lambda_{D}$). For each model, we report Utility (task-specific scores) and privacy metrics BDP ($\epsilon_\mu$) and RD. The RD metric is reported as max/avg, representing the worst-case and average-case divergence across all example pairs, respectively. Lower values are better for all privacy metrics ($\downarrow$).}
\label{tab:appendix_text_full_results_bert_large}
\centering
\small
\resizebox{\textwidth}{!}{%
\begin{tabular}{@{} l l c c c c c @{}}
\toprule
\textbf{Model} & \textbf{Metric} & \multicolumn{1}{c}{MRPC} & \multicolumn{1}{c}{STS-B} & \multicolumn{1}{c}{RTE} & \multicolumn{1}{c}{QNLI} & \multicolumn{1}{c}{SST-2} \\
\cmidrule(l){3-7}
& & Acc / F1 & Pearson / Spearman & Accuracy & Accuracy & Accuracy \\
\midrule
\multirow{2}{*}{BERT$_\text{Large}$ (Non-Private)} & Utility & 84.1 / 88.5 & 87.0 / 86.1 & 69.3 & 91.9 & 93.3 \\
& Privacy & - & - & - & - & - \\
\midrule
\multicolumn{7}{l}{\textbf{NVDP (Unconstrained)}} \\
\midrule
\multirow{3}{*}{$\lambda=1e-3$} & Utility & 84.6 / 88.8 & 85.1 / 84.9 & 69.2 & 92.2 & 93.0 \\
& BDP ($\epsilon_\mu$) $\downarrow$ & 10.70 & 20.27 & 10.80 & 10.86 & 13.68 \\
& RD (max/avg) $\downarrow$ & 1.555 / 0.150 & 2.993 / 0.259 & 0.846 / 0.071 & 1.277 / 0.101 & 0.911 / 0.129 \\
\cmidrule(l){2-7}
\multirow{3}{*}{$\lambda=1e-2$} & Utility & 85.0 / 89.2 & 82.1 / 82.4 & 51.9 & 92.0 & 94.3 \\
& BDP ($\epsilon_\mu$) $\downarrow$ & 10.60 & 20.27 & 10.69 & 10.73 & 10.48 \\
& RD (max/avg) $\downarrow$ & 0.611 / 0.087 & 2.516 / 0.129 & 0.168 / 0.048 & 0.559 / 0.099 & 0.529 / 0.059 \\
\cmidrule(l){2-7}
\multirow{3}{*}{$\lambda=1e-1$} & Utility & 66.5 / 75.9 & 75.9 / 82.4 & 50.3 & 50.5 & 50.6 \\
& BDP ($\epsilon_\mu$) $\downarrow$ & 10.54 & 18.62 & 10.59 & 10.47 & 10.47 \\
& RD (max/avg) $\downarrow$ & 0.026 / 0.006 & 0.412 / 0.037 & 0.018 / 0.004 & 0.017 / 0.003 & 0.030 / 0.008 \\
\cmidrule(l){2-7}
\multirow{3}{*}{$\lambda=1$} & Utility & 66.5 / 79.9 & 74.2 / 86.6 & 50.3 & 48.9 & 50.6 \\
& BDP ($\epsilon_\mu$) $\downarrow$ & 10.46 & 16.47 & 10.47 & 10.47 & 10.47 \\
& RD (max/avg) $\downarrow$ & 0.004 / 0.001 & 0.116 / 0.016 & 0.011 / 0.002 & 0.010 / 0.001 & 0.023 / 0.005 \\
\midrule
\multicolumn{7}{l}{\textbf{NVDP-Clipped (Ours)}} \\
\midrule
\multirow{3}{*}{$\lambda=1e-3$} & Utility & 85.4 / 89.5 & 86.6 / 85.5 & 69.7 & 92.2 & 93.2 \\
& BDP ($\epsilon_\mu$) $\downarrow$ & 10.54 & 16.10 & 10.73 & 10.86 & 10.65 \\
& RD (max/avg) $\downarrow$ & 0.313 / 0.029 & 0.264 / 0.028 & 0.109 / 0.015 & 1.015 / 0.054 & 0.153 / 0.013 \\
\cmidrule(l){2-7}
\multirow{3}{*}{$\lambda=1e-2$} & Utility & 83.3 / 87.9 & 86.8 / 85.8 & 68.7 & 92.0 & 93.6 \\
& BDP ($\epsilon_\mu$) $\downarrow$ & 10.53 & 15.93 & 10.58 & 10.70 & 10.47 \\
& RD (max/avg) $\downarrow$ & 2.891 / 0.084 & 0.384 / 0.046 & 0.104 / 0.019 & 0.270 / 0.042 & 0.122 / 0.018 \\
\cmidrule(l){2-7}
\multirow{3}{*}{$\lambda=1e-1$} & Utility & 82.1 / 87.2 & 86.6 / 85.6 & 51.3 & 48.8 & 50.2 \\
& BDP ($\epsilon_\mu$) $\downarrow$ & 10.51 & 15.71 & 10.51 & 10.47 & 10.46 \\
& RD (max/avg) $\downarrow$ & 0.020 / 0.006 & 0.296 / 0.030 & 0.004 / 0 & 0.001 / 0 & 0.008 / 0.002 \\
\cmidrule(l){2-7}
\multirow{3}{*}{$\lambda=1$} & Utility & 81.3 / 86.3 & 86.0 / 85.1 & 51.2 & 50.4 & 50.6 \\
& BDP ($\epsilon_\mu$) $\downarrow$ & 10.49 & 10.69 & 10.46 & 10.47 & 10.46 \\
& RD (max/avg) $\downarrow$ & 0.021 / 0.008 & 0.110 / 0.010 & 0.001 / 0 & 0.005 / 0 & 0.008 / 0.002 \\
\bottomrule
\end{tabular}
}
\end{table*}

\begin{table*}[t!]
\caption{Full experimental results for \textbf{RoBERTa-Base} on GLUE tasks, showing the complete privacy-utility trade-off across all tested KL-divergence regularization weights ($\lambda_{G}, \lambda_{D}$). For each model, we report Utility (task-specific scores) and privacy metrics BDP ($\epsilon_\mu$) and RD. The RD metric is reported as max/avg, representing the worst-case and average-case divergence across all example pairs, respectively. Lower values are better for all privacy metrics ($\downarrow$).}
\label{tab:appendix_text_full_results_roberta_base}
\centering
\small
\resizebox{\textwidth}{!}{%
\begin{tabular}{@{} l l c c c c c @{}}
\toprule
\textbf{Model} & \textbf{Metric} & \multicolumn{1}{c}{MRPC} & \multicolumn{1}{c}{STS-B} & \multicolumn{1}{c}{RTE} & \multicolumn{1}{c}{QNLI} & \multicolumn{1}{c}{SST-2} \\
\cmidrule(l){3-7}
& & Acc / F1 & Pearson / Spearman & Accuracy & Accuracy & Accuracy \\
\midrule
\multirow{2}{*}{RoBERTa$_\text{Base}$ (Non-Private)} & Utility & 86.9 / 90.3 & 89.2 / 88.6 & 73.5 & 91.8 & 94.0 \\
& Privacy & - & - & - & - & - \\
\midrule
\multicolumn{7}{l}{\textbf{NVDP (Unconstrained)}} \\
\midrule
\multirow{3}{*}{$\lambda=1e-3$} & Utility & 86.7 / 90.5 & 88.2 / 87.3 & 70.6 & 92.1 & 95.1 \\
& BDP ($\epsilon_\mu$) $\downarrow$ & 10.64 & 19.75 & 10.77 & 10.92 & 10.86 \\
& RD (max/avg) $\downarrow$ & 1.535 / 0.136 & 1.446 / 0.164 & 0.528 / 0.044 & 0.891 / 0.074 & 3.007 / 0.107 \\
\cmidrule(l){2-7}
\multirow{3}{*}{$\lambda=1e-2$} & Utility & 87.8 / 90.8 & 88.4 / 87.6 & 70.3 & 91.8 & 94.8 \\
& BDP ($\epsilon_\mu$) $\downarrow$ & 10.55 & 18.40 & 10.63 & 10.74 & 10.54 \\
& RD (max/avg) $\downarrow$ & 0.937 / 0.057 & 0.819 / 0.118 & 0.312 / 0.030 & 0.289 / 0.031 & 0.305 / 0.034 \\
\cmidrule(l){2-7}
\multirow{3}{*}{$\lambda=1e-1$} & Utility & 65.0 / 78.7 & 88.1 / 87.5 & 48.9 & 51.2 & 49.4 \\
& BDP ($\epsilon_\mu$) $\downarrow$ & 10.50 & 15.51 & 10.61 & 10.47 & 10.47 \\
& RD (max/avg) $\downarrow$ & 0.005 / 0.001 & 0.246 / 0.031 & 0.051 / 0.011 & 0.014 / 0.003 & 0.021 / 0.005 \\
\cmidrule(l){2-7}
\multirow{3}{*}{$\lambda=1$} & Utility & 66.4 / 79.8 & 87.8 / 87.2 & 49.6 & 51.1 & 49.3 \\
& BDP ($\epsilon_\mu$) $\downarrow$ & 10.47 & 15.50 & 10.47 & 10.47 & 10.47 \\
& RD (max/avg) $\downarrow$ & 0.003 / 0 & 0.237 / 0.027 & 0.005 / 0.001 & 0.013 / 0.003 & 0.019 / 0.004 \\
\midrule
\multicolumn{7}{l}{\textbf{NVDP-Clipped (Ours)}} \\
\midrule
\multirow{3}{*}{$\lambda=1e-3$} & Utility & 84.8 / 89.4 & 88.2 / 87.3 & 71.3 & 91.7 & 94.7 \\
& BDP ($\epsilon_\mu$) $\downarrow$ & 10.60 & 18.16 & 10.55 & 10.63 & 10.52 \\
& RD (max/avg) $\downarrow$ & 0.634 / 0.061 & 1.446 / 0.164 & 0.494 / 0.030 & 0.188 / 0.023 & 1.267 / 0.139 \\
\cmidrule(l){2-7}
\multirow{3}{*}{$\lambda=1e-2$} & Utility & 85.3 / 89.3 & 88.4 / 88.0 & 70.3 & 92.2 & 94.6 \\
& BDP ($\epsilon_\mu$) $\downarrow$ & 10.53 & 16.78 & 10.48 & 10.54 & 10.51 \\
& RD (max/avg) $\downarrow$ & 0.345 / 0.049 & 0.640 / 0.078 & 0.312 / 0.030 & 0.118 / 0.013 & 0.463 / 0.052 \\
\cmidrule(l){2-7}
\multirow{3}{*}{$\lambda=1e-1$} & Utility & 86.4 / 89.8 & 88.6 / 87.8 & 50.8 & 51.2 & 49.8 \\
& BDP ($\epsilon_\mu$) $\downarrow$ & 10.47 & 15.51 & 10.46 & 10.46 & 10.46 \\
& RD (max/avg) $\downarrow$ & 0.286 / 0.035 & 0.245 / 0.031 & 0.048 / 0.010 & 0.014 / 0.003 & 0.018 / 0.005 \\
\cmidrule(l){2-7}
\multirow{3}{*}{$\lambda=1$} & Utility & 66.4 / 79.8 & 87.8 / 87.2 & 49.6 & 49.3 & 50.1 \\
& BDP ($\epsilon_\mu$) $\downarrow$ & 10.46 & 15.4 & 10.46 & 10.46 & 10.46 \\
& RD (max/avg) $\downarrow$ & 0.003 / 0 & 0.237 / 0.027 & 0.005 / 0.001 & 0.012 / 0.003 & 0.013 / 0.003 \\
\bottomrule
\end{tabular}
}
\end{table*}
\clearpage

\end{document}

%% file: math_commands.tex
\usepackage{amsmath,amsfonts,bm}

\def\eqref#1{equation~\ref{#1}}

\def\1{\bm{1}}

\def\vx{{\bm{x}}}

\def\vz{{\bm{z}}}

\DeclareMathAlphabet{\mathsfit}{\encodingdefault}{\sfdefault}{m}{sl}
\SetMathAlphabet{\mathsfit}{bold}{\encodingdefault}{\sfdefault}{bx}{n}

